\documentclass[11pt,a4paper]{article}
\usepackage[T1]{fontenc}
\usepackage[utf8]{inputenc}
\usepackage[margin=22mm]{geometry}
\usepackage{newtxtext,newtxmath}
\usepackage{microtype}
\usepackage{graphicx}
\usepackage{booktabs,tabularx,array}
\usepackage{amsmath}
\usepackage{enumitem}
\usepackage{caption}
\usepackage[hidelinks]{hyperref}
\setlist{nosep,leftmargin=*}
\newcommand{\NULL}{\texttt{NULL}}
\newcommand{\START}{\texttt{START}}
\newcommand{\END}{\texttt{END}}
\newcommand{\UNKNOWN}{\texttt{UNKNOWN}}
\newcommand{\m}{\,\mathrm{m}}
\newcommand{\doi}[1]{\href{https://doi.org/#1}{doi: \nolinkurl{#1}}}
\newcolumntype{Y}{>{\raggedright\arraybackslash}X}
\title{Waggle Dance Inspired Motion Communication\\for Multiple UAVs in MuJoCo}
\author{Zhang Nengbo\\[3pt]
\small School of Aerospace Engineering, Engineering Campus\\
\small Universiti Sains Malaysia, 14300 Nibong Tebal, Pulau Pinang, Malaysia\\
\small\texttt{zhangnb@student.usm.my}}
\date{}
\begin{document}
\maketitle
\begin{abstract}
The honeybee waggle dance motivates a communication mechanism in which one agent's movement conveys spatial information that guides other agents' actions. This paper presents a MuJoCo system that extends the point-to-point motion communication setting of MoCom to one performer and multiple observers. A performer broadcasts a six-bit navigation payload using four flight primitives and explicit null signals. Each of one to five observers processes its own onboard RGB images, extracts optical-flow trajectories, recognizes symbols, parses the message, and starts navigation only after confirming its own complete frame. Reception states and execution triggers are separate across observers, while simulation control and safety checks use shared ground truth. With stationary observers, 25 Hz image input, and ideal state-feedback control, a fixed standard suite yielded 44 correct complete messages from 53 receiver exposures across 17 nominal broadcasts; 13 broadcasts passed all group-level decoding and execution checks. Three additional no-message or input-fault controls met their expected outcomes. A separately reported supplemental suite, using the same frozen code at the default geometry, achieved 14 successful receiver exposures across three broadcasts. Near-range and wide-angle configurations exposed tracking and recognition failures, while unsuccessful receivers remained stationary. These finite simulation results support the feasibility of a waggle-dance-inspired broadcast-to-action mechanism under the tested conditions and identify the present perceptual and protocol limits.
\end{abstract}

\section{Introduction}
The waggle dance is relevant to engineered communication because a receiver can use information expressed through another individual's movement to guide subsequent action. Radar tracking of recruited honeybees provides direct evidence connecting dance communication with spatial flight behavior~\cite{riley2005}. For micro aerial vehicles (MAVs), this suggests a testable system abstraction: one vehicle deliberately moves to encode a message, nearby vehicles observe that movement, and each receiver translates the recovered message into an action. The abstraction concerns the functional relationship between signaling, reception, and execution; a discrete flight codebook need not reproduce the physical trajectory of a bee's dance.

Motion-based robot communication already has relevant precedents. MRoCS explored passive action recognition for bee-inspired communication in simulated and physical robot settings, including aerial robots~\cite{das2016}. MoCom investigated inter-MAV visual communication using event vision and spiking neural networks, including null intervals and navigation messages followed by execution~\cite{mocom2025}. These studies motivate, rather than eliminate, the need to verify a multi-receiver system. A successful point-to-point demonstration does not establish that several observers, viewing the same performer from different positions, will each recover the message or start the intended action from their own visual evidence.

This paper addresses three system questions. First, can a single motion broadcast be decoded separately by multiple observers? Second, can each observer trigger the corresponding action from its own complete message without a group-wide release signal? Third, how do receiver count, viewing geometry, and a local input fault affect completion of the communication-to-action chain? The implementation uses ordinary RGB cameras and sparse optical flow as a transparent first receiver, allowing individual trajectories, symbols, frame decisions, and navigation states to be inspected.

The contribution is a simulation system and a bounded mechanism demonstration. Specifically, the system instantiates one to five independent receiver state chains; connects visually confirmed frame completion to per-observer navigation; and records both successful and failed configurations in a frozen-code validation matrix. Motion communication, null signaling, and navigation payload fields are inherited ideas. The present work does not claim the first bee-inspired aerial communication system or a new optical-flow recognition algorithm. Instead, it makes the extension from a point-to-point link to multiple independently acting receivers concrete and auditable.

\section{Related work and biological interpretation}
\paragraph{Reading and reproducing dance communication.}
In addition to behavioral evidence from recruited bees~\cite{riley2005}, prior work has addressed how dance movements can be extracted from video. Landgraf and Rojas studied honeybee dance tracking using sparse optical-flow fields~\cite{landgraf2007}; Wario et al. developed an automatic detector and decoder of honeybee waggle dances~\cite{wario2017}. RoboBee explored robotic reproduction of dance communication and recruitment in a biological colony~\cite{landgraf2018}. These works distinguish observing dance motion, interpreting its spatial content, and influencing a receiver's behavior. They also establish that optical flow applied to dance analysis is not itself a new concept introduced here.

\paragraph{From motion recognition to multiple executing receivers.}
MRoCS used recognized robot actions as communication signals~\cite{das2016}. MoCom defined a four-primitive visual codebook and an event-based processing pipeline for inter-MAV communication~\cite{mocom2025}. The present receiver instead uses RGB feature tracking based on Lucas--Kanade optical flow and trackable image features~\cite{lucas1981,shi1994}. This change is an implementation choice for an inspectable simulator prototype, not an evaluated improvement over the MoCom event-camera pipeline. The distinction studied here is at the system level: every observer has its own images, null confirmation, frame parser, and navigation trigger.

The biological correspondence is deliberately limited. The performer acts as a source of spatial instructions, and multiple receivers can use the same observed signal to guide motion. The current message means the same displacement in a pre-agreed coordinate frame for every receiver. It does not encode a learned food source, implement competing recruitment, or reproduce colony-level foraging decisions. Demonstrating those behaviors would require further models and experiments.

\section{System architecture}
\subsection{Scene, sensors, and information boundaries}
The scene contains one performer and $N\in\{1,\ldots,5\}$ observers in the MuJoCo rigid-body simulator~\cite{todorov2012}. Vehicle geometry and inertial resources are based on the Crazyflie 2 model in MuJoCo Menagerie~\cite{menagerie}. Observers hover and navigate at $z=1.3\m$. The performer's hover and horizontal signaling share this nominal height; vertical primitives may depart from it. Each observer has a virtual RGB camera with resolution $640\times480$ and a vertical field of view of $60^\circ$. Its initial yaw faces the known performance area. Observer ego motion is disabled during reception.

\begin{figure}[tbp]
\centering
\includegraphics[width=\linewidth]{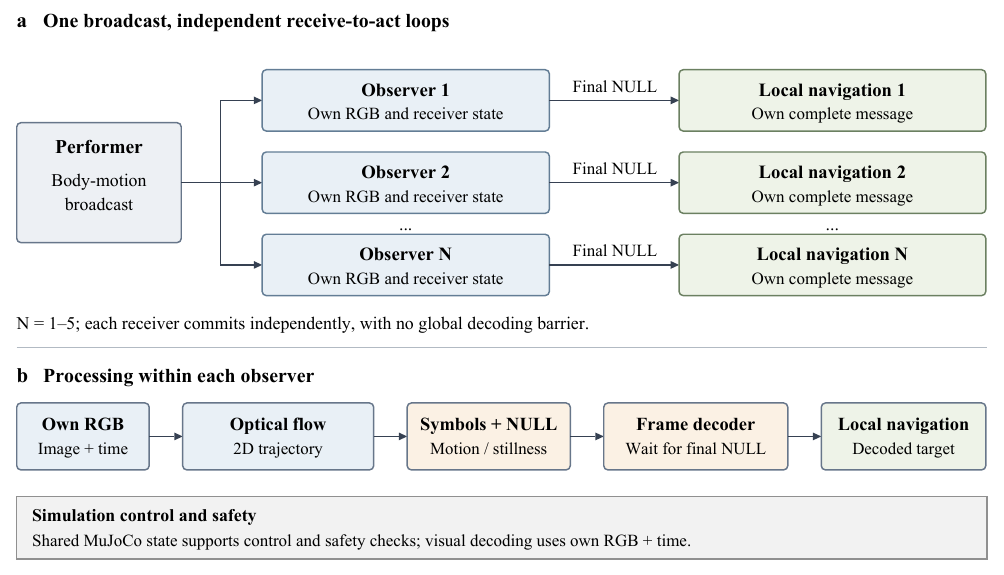}
\caption{One performer broadcasts motion to multiple observers. Each receiver has its own visual processing and frame state; its final null confirmation releases only its own decoded command. Simulator state feedback and safety checks are shared infrastructure, not evidence supplied to the visual decoder.}
\label{fig:system}
\end{figure}

Each observer maintains a separate tracker, motion segmenter, null timer, message decoder, and navigator (Fig.~\ref{fig:system}). The perception interface accepts only that observer's RGB image and timestamp. It does not receive the transmitted bits, current transmitter phase, simulated trajectory, truth segmentation, or another receiver's prediction. Initial acquisition searches near the image center, which is a declared initialization prior rather than unrestricted target detection.

All nodes share a simulation clock, and their cameras are rendered and processed sequentially at each common sampling time. Thus, independence refers to the inputs and state transitions of the receiver chains. It does not mean separate onboard computers, asynchronous physical camera hardware, or fully distributed control. Simulator truth remains available to the controllers, safety checks, and evaluator; these uses are separate from symbol recognition.

\subsection{Motion alphabet and message framing}
The motion alphabet contains four primitives (Table~\ref{tab:alphabet}). A six-bit payload follows the direction--heading--distance organization used by MoCom~\cite{mocom2025}: $d\,|\,hhh\,|\,rr$, where $d$ is one direction bit, $h$ is a three-bit heading integer, and $r$ is a two-bit distance integer. Null signals separate adjacent motion symbols and confirm the frame boundaries. They are not payload bits.

\begin{table}[tbp]
\centering\small
\caption{Visual symbols and their protocol roles.}\label{tab:alphabet}
\begin{tabularx}{\linewidth}{@{}Y l Y@{}}
\toprule
Observed behavior & Symbol & Protocol role\\
\midrule
Vertical back-and-forth motion & \START & Begin a data frame\\
V-shaped motion & \texttt{0} & Binary payload bit\\
Inverted V-shaped motion & \texttt{1} & Binary payload bit\\
Horizontal back-and-forth motion & \END & Await final confirmation\\
Reliable stillness near the image anchor & \NULL & Separate symbols and confirm boundaries\\
\bottomrule
\end{tabularx}
\end{table}

For example, payload \texttt{000010} produces the complete visual event sequence
\begin{center}\small\ttfamily
NULL START NULL 0 NULL 0 NULL 0 NULL\\
0 NULL 1 NULL 0 NULL END NULL.
\end{center}
There are eight motion symbols and nine null events, for 17 events in total. The parser requires a valid start, all six payload bits with separators, an end symbol, and the final null. An \END{} event alone enters a waiting state; it does not submit a navigation command. Only that receiver's final visually confirmed \NULL{} completes the frame. Missing starts, unknown gestures, incomplete separators, and truncated frames cannot be submitted as valid commands under this parser. The construction explicitly separates partial recognition from acceptance of an executable message.

\subsection{RGB optical flow and trajectory recognition}
The receiver tracks image features using pyramidal Lucas--Kanade flow, with feature selection based on trackable local image structure~\cite{lucas1981,shi1994}. Forward--backward consistency and support checks reject unreliable tracks. Robust aggregation of the surviving signed displacements produces a two-dimensional motion trajectory. The first version assumes a stationary observing camera; it does not remove camera-induced flow.

Motion begins when the image speed exceeds $3.5$ pixels/s for $0.08$ s and ends after speed stays below $2.5$ pixels/s for $0.68$ s. Each extracted trajectory is resampled at 80 arc-length positions, translated to its starting point, and divided by its largest axis-aligned span. The normalized path is compared against ordered templates. Rotation is not normalized away because the vertical and horizontal directions carry distinct symbol meanings.

Acceptance requires a normalized template root-mean-square error below $0.17$, a best-to-second-best error margin above $0.075$, a normalized closure error below $0.18$, and an observed span of at least 28 pixels. Otherwise the receiver returns \UNKNOWN{} rather than forcing a bit decision. The template score is a geometric agreement measure, not a calibrated class probability. These thresholds were held fixed during the reported validation suites.

\subsection{Visual null confirmation}
Null detection requires a stationary target that has returned near its initial image anchor. With no active gesture, the center must remain within 15 pixels of that anchor, tracking quality must be at least $0.2$, and speed must stay below $2.5$ pixels/s continuously for 3 s. Motion events, tracking loss, substantial frame gaps, and departures from the anchor reset the timer. A quiet segment produces at most one null event.

The transmitter schedules a 2 s explicit hover gap. Together with stillness at the gesture ends and the following gesture's lead-in, this gives the receiver time to satisfy its 3 s visual test. The receiver never reads the transmitter's gap timer. Null intervals were already part of MoCom~\cite{mocom2025}; the present system specifies their RGB evidence and state transitions. Reliable stillness can reject the tested black-input fault, but it does not establish sensor liveness against repeated static images.

\subsection{Per-observer navigation}
A receiver's complete message is passed directly to its own navigator, without waiting for all observers to succeed. Let $\mathbf p_i(t_i)$ denote observer $i$'s position when it accepts its complete frame. All receivers pre-agree that command heading zero is world $+Y$, with counterclockwise headings in a top view. Using heading step $\alpha=\pi/4$ rad and distance unit $u=0.2\m$, the target is
\begin{align}
\theta &= \frac{\pi}{2}+h\alpha,\label{eq:heading}\\
\mathbf p_i^* &= \mathbf p_i(t_i)+(-1)^d r u
\begin{bmatrix}\cos\theta&\sin\theta&0\end{bmatrix}^{\!\top}.
\label{eq:target}
\end{align}
The target height is fixed at $1.3\m$. Payloads \texttt{000010}, \texttt{000110}, \texttt{001010}, and \texttt{100010} therefore command displacements of approximately $(0,0.4,0)$, $(-0.283,0.283,0)$, $(-0.4,0,0)$, and $(0,-0.4,0)\m$, respectively. These numeric units and the world-axis convention are explicit simulator choices, not a claim to reproduce all coordinates from the earlier MoCom figures. Camera yaw does not define the command heading.

The common broadcast means an equal displacement, not convergence to one absolute target. After each reset, an observer executes at most one accepted command; its recognition stops during navigation while camera rendering continues. Smooth reference motion is followed using ideal forces and torques with state feedback through MuJoCo dynamics, rather than teleporting the body to the destination. Arrival requires position error below 1.5 cm and speed below 2.5 cm/s for 0.4 s. Safety checks inspect true positions and planned paths and conservatively reject unsafe spacing; they do not implement distributed onboard avoidance or replanning.

\section{Evaluation protocol}
\subsection{Fixed standard suite}
The standard suite contained 17 nominal complete broadcasts and three separate controls. Nominal means that a complete message was intentionally transmitted; it does not imply favorable viewing conditions. The configurations varied observer count from one to five, arc versus line layouts, distance parameters of 1.0, 1.3, and 1.6 m, angular half-spans of $15^\circ$, $30^\circ$, $35^\circ$, $40^\circ$, and $45^\circ$, and the four payloads above. Selected runs added initialization jitter of up to $\pm0.01\m$ on each horizontal axis with seeds 0 or 1. This was a finite configuration matrix, not the full Cartesian product of factors. Jitter affected initial placement only and did not represent image noise or sustained ego motion.

In an arc, the distance parameter is the radius; in a line it is the longitudinal separation. The default was three observers, an arc at 1.3 m with a $30^\circ$ half-span, no jitter, and payload \texttt{000010}. A single observer is directly in front of the performer, so its spread parameter has no geometric effect. Every receiver processed images at 25 Hz. Full configurations and outcomes are listed in Table~\ref{tab:results}.

\subsection{Metrics and provenance checks}
\emph{Complete-message correctness} counts a receiver exposure as correct only if the entire six-bit payload is recovered. The descriptive receiver proportion divides correct exposures by all configured exposures in nominal broadcasts. \emph{Whole-group completion} requires every observer in one broadcast to decode correctly, start from its own accepted message, and satisfy the navigation and hold checks without premature movement or contact. Failed receivers remain in the denominator. No-message and fault controls are reported separately from communication success.

For successful receivers, the evaluator verifies the full 17-event sequence and nine null events, the relation between final-null and navigation-start timestamps, and at least 2 s of holding after arrival. Expected displacement is calculated independently from the test payload and the fixed world-$+Y$ convention; the evaluator does not read the tested navigator's or camera's heading to construct its answer. Input-boundary inspection and runtime message provenance checks accompany these behavioral tests. This is a bounded audit, not a formal proof of information-flow isolation.

\subsection{Controls and supplemental suite}
Three controls test distinct failure paths. In the no-send control, all vehicles remain stationary for 10 s: an initial null is allowed, but no complete message or navigation may occur. In the local input-fault control, observer 2 receives entirely black images while observers 1 and 3 retain their own images. This models unavailable visual input, not physical occlusion geometry. In the interruption control, transmission stops at 26 s and observation continues until 60 s; the incomplete frame must not start navigation.

After the standard suite, three supplemental broadcasts used the same frozen source at the common default distance and half-span: four-observer forward, five-observer forward, and five-observer backward motion. These post-hoc checks ask whether higher receiver counts work at the common default geometry. They are reported separately, without replacing standard failures or pooling denominators. Source hashes were unchanged within and across the suites.

A broadcast is the experimental unit: observers in the same broadcast share a performer and scene, so their outcomes may be correlated. Neither video frames nor simultaneous observers are treated as independent random repetitions. The reported proportions describe this deterministic engineering matrix and are not estimates of population reliability.

\section{Results}
\subsection{Broadcast reception and action completion}
The standard nominal broadcasts produced 44 correct complete messages among 53 receiver exposures (83.02\%). Thirteen of 17 broadcasts passed all whole-group decoding and execution checks (76.47\%). At the default three-observer geometry, all four payloads were recovered and executed by all observers. The other successful standard configurations included one and two observers, a four-observer arc with a $35^\circ$ half-span, the line layout, the 1.6 m range, the narrow arc, and the tested initialization-jitter cases (Table~\ref{tab:results}).

\begin{table}[tbp]
\centering\small
\caption{All nominal broadcasts. Unless stated otherwise: arc, 1.3 m, half-span $30^\circ$, no jitter, seed 0, and payload \texttt{000010}. $N$ counts observers only. ``Group'' means all decoding and navigation checks passed. ``Correct'' counts complete receiver messages. Initial jitter $j=0.01$ denotes independent bounded XY perturbations. Supplemental trials use the same frozen code and are not pooled with the standard suite.}
\label{tab:results}
\setlength{\tabcolsep}{4pt}
\begin{tabularx}{\linewidth}{@{}l c Y c c@{}}
\toprule
Suite & $N$ & Configuration change & Correct & Group\\
\midrule
Standard & 3 & Default, forward & 3/3 & Pass\\
& 3 & \texttt{000110}, heading $45^\circ$ & 3/3 & Pass\\
& 3 & \texttt{001010}, heading $90^\circ$ & 3/3 & Pass\\
& 3 & \texttt{100010}, backward & 3/3 & Pass\\
& 1 & One observer, front view & 1/1 & Pass\\
& 2 & Two observers & 2/2 & Pass\\
& 4 & Half-span $35^\circ$ & 4/4 & Pass\\
& 5 & Half-span $40^\circ$, forward & 3/5 & Fail\\
& 3 & Line layout & 3/3 & Pass\\
& 3 & Distance 1.0 m & 0/3 & Fail\\
& 3 & Distance 1.6 m & 3/3 & Pass\\
& 3 & Half-span $15^\circ$ & 3/3 & Pass\\
& 3 & Half-span $45^\circ$ & 1/3 & Fail\\
& 3 & $j=0.01$, seed 0 & 3/3 & Pass\\
& 3 & $j=0.01$, seed 1 & 3/3 & Pass\\
& 3 & Line, $j=0.01$, seed 1, \texttt{000110} & 3/3 & Pass\\
& 5 & Half-span $40^\circ$, \texttt{100010} & 3/5 & Fail\\
\midrule
\multicolumn{3}{@{}l}{Standard total: 17 broadcasts} & 44/53 & 13/17\\
\midrule
Supplement & 4 & Default geometry, forward & 4/4 & Pass\\
& 5 & Default geometry, forward & 5/5 & Pass\\
& 5 & Default geometry, \texttt{100010} & 5/5 & Pass\\
\midrule
\multicolumn{3}{@{}l}{Supplemental total: 3 broadcasts} & 14/14 & 3/3\\
\bottomrule
\end{tabularx}
\end{table}

The four failed standard configurations were not crashes or discarded trials. Both five-observer trials with a $40^\circ$ half-span succeeded at the three inner observers but failed at the two outer observers. The three-observer 1.0 m trial yielded no complete messages, and the $45^\circ$ half-span trial succeeded only at the middle observer. All nine unsuccessful receiver exposures remained waiting without moving.

The supplemental suite achieved 14 correct receiver messages and arrivals in three of three broadcasts. Together with the standard suite, this provides at least one complete closed-loop record for each count from one to five at the common default parameters, with the single-observer geometric exception noted above. It establishes demonstrated operation at those tested configurations, not a universal five-receiver reliability guarantee.

\subsection{Local faults and independent release}
All three controls met their respective expectations (Table~\ref{tab:controls}). In particular, when observer 2 received black images, observers 1 and 3 decoded and arrived using their own receivers, while observer 2 neither completed a frame nor moved. Thus, successful peers did not supply a replacement result to the failed node, and the failed node did not create a group-wide execution barrier in this test.

\begin{table}[tbp]
\centering\small
\caption{Controls, reported separately from nominal communication proportions.}
\label{tab:controls}
\begin{tabularx}{\linewidth}{@{}p{0.28\linewidth} Y@{}}
\toprule
Control & Observed outcome\\
\midrule
Static, no send, 10 s & No complete message or navigation at any observer.\\
Black input at observer 2 & Observers 1 and 3 decoded and arrived; observer 2 waited without moving.\\
Transmission stopped at 26 s & No complete message or navigation while observation continued to 60 s.\\
\bottomrule
\end{tabularx}
\end{table}

Per-observer event logs show navigation starting at the same timestamp as that observer's final null acceptance. In the line-layout heading trial with initialization jitter, observer 1 accepted and started at 95.60 s, whereas observers 2 and 3 started at 95.64 s. This 0.04 s difference, together with the recorded per-node message provenance, is consistent with local release rather than a shared decode-completion barrier. All times are simulation times.

In the default three-observer forward example, \END{} was recognized at approximately 92.64 s, the final null at 95.64 s, and arrival at 98.784 s. The interval between recognizing \END{} and accepting the message is therefore an intentional confirmation stage. A full message carries only six payload bits, and this first implementation has a long signaling latency. No throughput advantage is claimed.

\subsection{Perceptual failure modes and physical checks}
Failure traces locate the main problems in visual reception. At 1.0 m, the \START{} segment encountered KLT support or forward--backward consistency failures. Subsequent recovery of some data bits could not repair a frame without its required start. At the outer views of the $40^\circ$ configuration, V-shaped trajectories also exceeded the fixed template error threshold: representative errors were 0.1727 and 0.1758, above 0.17. The outer observers at $45^\circ$ did not form complete messages. Recognition thresholds were not relaxed for these cases.

No body contact was recorded in any of the 23 standard and supplemental cases. Observers remained stationary before complete-message acceptance, and every successful receiver met the specified arrival and holding checks. These are properties of the ideal force/torque controller and truth feedback in this simulator. Extremely small simulated position errors should not be interpreted as physical Crazyflie positioning accuracy. The finite viewing configurations identify observed failure cases but do not determine a continuous critical distance or angular boundary.

\section{Discussion}
\subsection{What the mechanism demonstration establishes}
The central result is a working chain from one visible motion message to several independently decoded and executed commands. Compared with a point-to-point link, the multi-receiver system exposes different image projections, local failures, and unequal frame-completion times. Independent parsing and final-null release preserve these differences rather than hiding them behind a single shared receiver result. The observations answer the system questions at the level of the tested prototype: a broadcast can lead to several actions, local frame completion can trigger each action, and viewing conditions can prevent a subset of receivers from participating.

Successful default-geometry trials and failed outer-view trials indicate that receiver count alone is insufficient to describe scalability. Each added receiver must also occupy a usable viewpoint. However, receiver number and angle were not fully separated in the standard matrix, so these results do not isolate a causal effect of group size. A larger factorial evaluation would be needed to distinguish count, distance, and viewpoint effects quantitatively.

The connection to waggle dance is functional: bodily motion conveys spatial content and changes receiver behavior. The current shared coordinate convention supplies the meaning of a displacement in advance. There is no coordinate-frame negotiation, common absolute destination, heterogeneous task assignment, recruitment competition, multi-performer arbitration, relay communication, or collective consensus. Those are possible extensions of the research program, not demonstrated capabilities of this implementation.

\subsection{Limitations and next simulation experiments}
The receiver assumes known initial target direction and a stationary camera. Geometry templates are sensitive to projection changes and tracking failures, as the reported cases show. Learned recognition, event-camera processing, and ego-motion compensation have not been compared here. Reliable null detection also cannot generally distinguish genuine hovering from repeated static camera frames. It is neither a sensor-liveness guarantee nor a payload checksum. There is no acknowledgement, retransmission, error-correcting code, or continuous multi-packet evaluation.

The physics layer uses ideal applied forces and torques with true-state feedback. Motor dynamics, rotor aerodynamics, downwash, realistic onboard estimation errors, and fully distributed collision avoidance are outside the evaluated model. Similarly, a single software process with separate receiver states is not evidence of real-time deployment on multiple onboard computers. The study makes no hardware or sim-to-real claim.

The evaluation is a finite deterministic engineering suite, without large randomized repetition, an independent generalization test set, baseline comparisons, or ablations. Its percentages are descriptive. Subsequent simulation experiments should separate count, distance, and angle; repeat trials under controlled image noise and initialization distributions; and compare receiver variants. Ablations of null separators and final confirmation would help quantify their contribution. Protocol duration and meaningful payload capacity also need systematic measurement before the mechanism can be treated as an efficient operational communication channel.

\subsection{Reproducibility and system inspection}
The implementation records configurations, source SHA-256 hashes, per-observer images and events, complete-message outputs, navigation states, and scoring results. The standard and supplemental outputs are kept separately, including the failed cases. The interface provides a selected observer view, a performer view, a vertical top-down view, and a third-party overview. Recorded composites can include every observer together with the three shared views; optional raw RGB capture permits comparison with the actual receiver input.

These records allow perception failure, frame rejection, and navigation behavior to be inspected as separate stages. The source snapshots remained identical across the reported suites. The present paper reports the system and its finite validation evidence; it does not claim an already public external code or data repository.

\section{Conclusion}
This paper presents a waggle-dance-inspired motion communication system linking one performer to multiple independently receiving and acting observers in MuJoCo. Each receiver obtains its own visual evidence, confirms a complete six-bit message, and starts its own navigation. Closed-loop operation was recorded for one to five observers at the common default settings, while near-range and wide-angle tests exposed perceptual failures without inducing motion at undecoded nodes. The system provides a concrete simulation basis for extending MoCom from point-to-point communication to multi-receiver signal-to-action coordination, with further work required on visual robustness, protocol efficiency, and controlled scale evaluation.

\bibliographystyle{unsrt}
\bibliography{references}
\end{document}